\documentclass[num-refs]{nbdt-article}

\usepackage{natbib}
\usepackage{siunitx}
\usepackage{ccfonts}
\usepackage{multirow}
\usepackage{graphicx} 
\papertype{Original Article}

\title{WhACC: Whisker Automatic Contact Classifier with Expert Human-Level Performance}

\abbrevs{
AUC, Area Under the Curve;
CNN, convolutional neural network;
fps, Frames Per Second;
PSTH, Peri-Stimulus Time Histogram;
SD, Standard Deviation;
TC-error, Touch Count Error;
WhACC, Whisker Automatic Contact Classifier
}

\author[1,2,3]{Phillip Maire}
\author[1,2]{Samson G. King}
\author[1,2]{Jonathan Andrew Cheung}
\author[1] {Stefanie Walker}
\author[1,3]{Samuel Andrew Hires}

\affil[1]{Department of Biological Sciences, Section of Neurobiology, University of Southern California, Los Angeles, California, United States of America.}
\affil[2]{Neuroscience Graduate Program, University of Southern California, Los Angeles, California, United States of America.}
\affil[3]{Lead Contacts}

\corraddress{Phillip Maire \& Samuel Andrew Hires, \\Section of Neurobiology, \\Department of Biological Sciences, \\University of Southern California, \\Los Angeles, California, 90089}
\corremail{Phillip.Maire@gmail.com \& shires@usc.edu}

\fundinginfo{NIH National Institute of Neurological Disorders and Stroke, Grant Number R01NS102808}

\runningauthor{Phillip Maire et al.}

\begin{document}

\maketitle

\begin{abstract}
\normalsize
The rodent vibrissal system is pivotal in advancing neuroscience research, particularly for studies of cortical plasticity, learning, decision-making, sensory encoding, and sensorimotor integration. Despite the advantages, curating touch events is labor intensive and often requires >3 hours per million video frames, even after leveraging automated tools like the Janelia Whisker Tracker. We address this limitation by introducing Whisker Automatic Contact Classifier (WhACC), a python package designed to identify touch periods from high-speed videos of head-fixed behaving rodents with human-level performance. WhACC leverages ResNet50V2 for feature extraction, combined with LightGBM for Classification. Performance is assessed against three expert human curators on over one million frames. Pairwise touch classification agreement on 99.5\% of video frames, equal to between-human agreement. Finally, we offer a custom retraining interface to allow model customization on a small subset of data, which was validated on four million frames across 16 single-unit electrophysiology recordings. Including this retraining step, we reduce human hours required to curate a 100 million frame dataset from $\sim$333 hours to $\sim$6 hours.

\end{abstract} 

\section{Introduction}
Quantitative analysis of behavior is an essential method in systems neuroscience research. High speed video is a commonly used format for recording behavior. The resulting datasets are often large and, if they require manual curation, the analytical time cost is large as well. One such pain point is found within investigations using the rodent whisker system, a common model for investigating neural representations of tactile perception and sensorimotor integration. As the stimuli used in this field have evolved from passive whisker deflections in anesthetized rodents to active whisker touches during behavior, new methods for quantification of these stimuli are required. Here we address a specific and challenging problem, fully automated whisker touch detection from videography.

Rodents are highly tactile creatures that sweep an array of whiskers forward and back during many behaviors, including locomotion \cite{sofroniew_whisking_2015,grant_whisker_2018}, social interaction \cite{rao_vocalization-whisking_2014} and object investigation \cite{cheung_sensorimotor_2019, cheung_independent_2020, kim_behavioral_2020}. When whiskers touch something, the resulting forces drive mechanotransduction in the follicle and propagation of neural activity to higher order brain regions. The neural circuitry displays remarkable temporal precision \cite{jadhav_sparse_2009, andrew_hires_low-noise_2015, bale_organization_2018}, so identifying the time of touch with millisecond resolution is crucial for investigations of tactile processing. Demonstrating this importance, automated classification programs such as Biotact \cite{Lepora_NaiveBayes_2011}, Janelia Whisker Tracker \cite{clack_automated_2012}, and related tools \cite{knutsen_tracking_2005, voigts_unsupervised_2008, perkon_unsupervised_2011, towal_morphology_2011, betting_whiskeras:_2020}, have been developed to identify when, where, and how hard whiskers are touching objects in head-fixed and freely moving as well as full-field, reduced whisker and single whisker paradigms.

These whisker tracking methods provide faster touch classification than hand scoring, but none achieve maximally accurate touch classification without a second stage of manual curation of classification results. In the simple case of the Janelia Whisker Tracker applied to head-fixed, single whisker touch classification, our second stage manual curation process requires approximately 3 hours and 20 minutes per million video frames to complete. To alleviate this time burden, we developed, trained and validated a hybrid convolution neural network and gradient boosted machine model to accurately identify single-whisker touch directly from video frames in a fully automated manner. 

\subsection{Design and Implementation}
The overall design goal was to rapidly classify high-speed video of head-fixed object localization in mice and accurately identify periods when a whisker was touching a presented object. The imaging viewpoint was overhead from a single camera with the whisker pad, whisker and object (a thin pole) backlit from a diffuse infrared light-emitting diode (Figure 1A). The resulting touch labels from analysis can then be used to characterize electrophysiological responses of neurons to touch (Fig 1A inset).

A full frame video is useful for determining various kinematic features of whisker motion and deformation, such as velocity and curvature from bending. However, the most relevant information for touch identification is found in a small window of pixels around the object. Additionally, small images reduce data size and speeds up the training process. Therefore, video input to the model was cropped by extraction of a 61x61 pixel window centered on the touched object across all video frames (Figure 1A, red box). This eliminated extraneous and idiosyncratic image data (e.g., fur or stubs of other whiskers) that could impede model performance by fitting to unreliable features. Independently centering the object on each frame also accommodates the potential translation of the touched object across the field of view (e.g., when the object is being presented, withdrawn, or is vibrating). 

When discriminating challenging touches, human curators often scroll forward and back between a few frames to infer when touch onset or offset occurs by the change in whisker displacement between frames. To provide the classifier with this temporal information, we overlaid three consecutive grayscale frames into three color channels (cyan, magenta, and yellow), such that the label at time t had access to the image at time t, and lag images at time t-1, and t-2 (Figure 1B, left). To improve the generality of our training set to different imaging conditions and orientations, we augmented images with rotation, noising, brightness and magnification adjustments (Figure 1B, right). These sets of images were used as inputs to a variety of convolutional neural networks (CNNs) (Figure 1C).

We extracted all 2,048 available features from the penultimate layer of the best performing model and liberally engineered additional features based on the standard deviation, smoothing, shifting and discrete difference of different window/step sizes, resulting in 40 additional sets of 2048 features, for a total of 83,968 features (Methods). Additionally we calculated standard deviation of each of the 41 feature sets across feature space for a total of 84,009 features (Figure 1D, left). To reduce the computer memory demands and reduce model complexity, we iteratively selected smaller subsets of the most informative features for touch classification using an ensemble of LightGBM models, based on gain and split importance measures. Once further feature removal degraded mean model performance, we had a total of 2,105 features remaining (Figure 1D, right).

\begin{figure}[t]
\centering
\includegraphics[width=\textwidth] {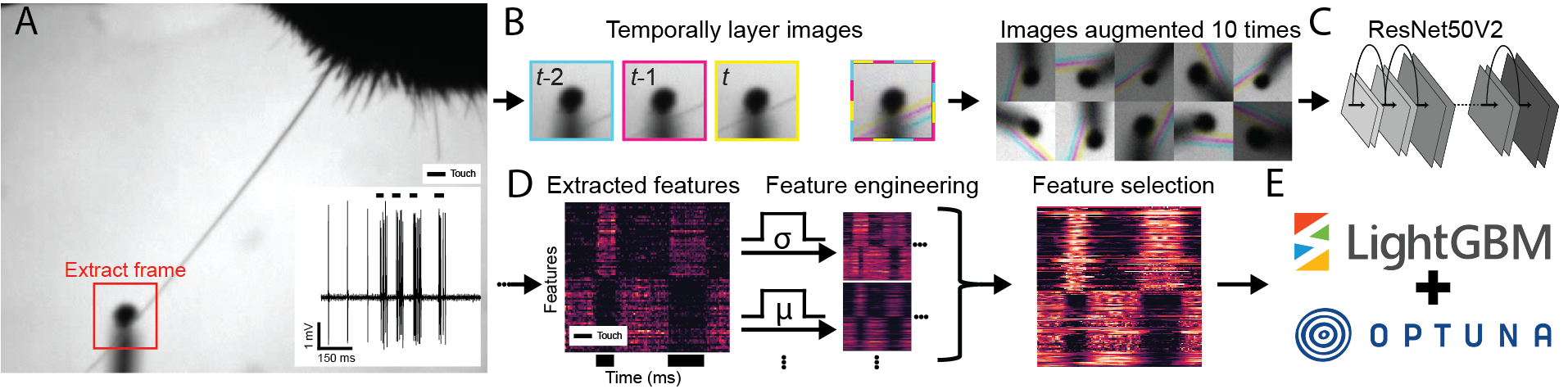}
\caption{\textbf{Flow diagram of WhACC video pre-processing and design implementation.} A) Sample touch frame from high-speed (1,000 fps) video and extracted object-centered window for CNN input (red box) and corresponding spike train from a touch responsive neuron (inset) B) Three consecutive extracted frames combined into three color channels (left) and example augmented images (right). C) Representation of ResNet50V2 model used to extract features. D) Demonstrative sample of features extracted from ResNet50V2 (left), representation of feature engineering (center) and feature selection (right) for final WhACC model. E) Final WhACC model was trained using LightGBM with Optuna to achieve the best performance. 
}
\end{figure}

\begin{figure}[bt]
\centering
\includegraphics[width=\textwidth] {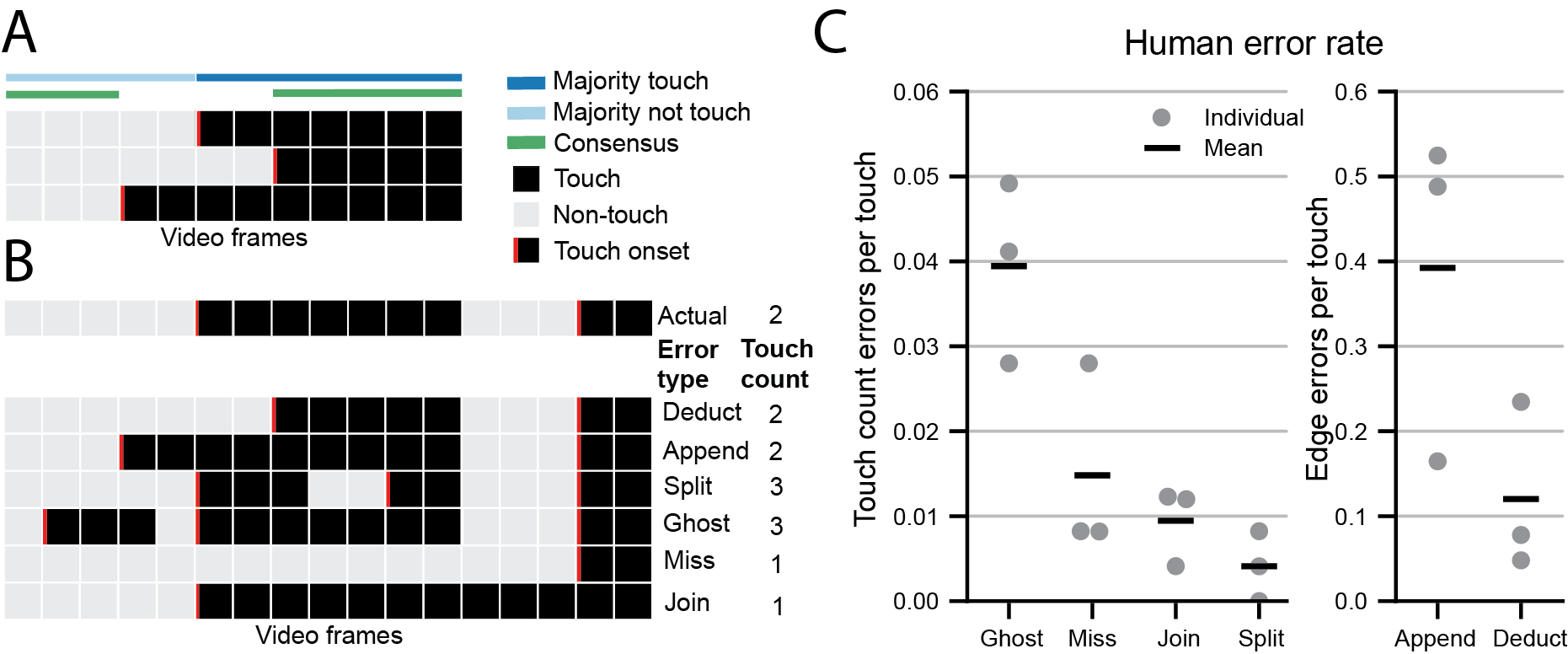}
\caption{\textbf{Touch frame scoring and variation in human curation.} A) Example of disparity between three human curators. Majority touch (dark blue), majority non-touch (light blue) were used for training models. Consensus frames (green) were used when evaluating curator versus paired consensus in C. B) Example of scored touch array (human majority) and the corresponding edge errors (deduct and append), and touch count errors (split, ghost, miss and join). C) Individual and mean error rate for each human curator compared against the consensus of the other two curators for touch count errors (left) and edge errors (right).}
\end{figure}

\subsection{Establishing touch ground truth and error metrics}
Training a supervised classifier requires a set of inputs and a corresponding set of accurately classified labels. However, there is no independent ground truth for whether a whisker is touching the object or not on each video frame. Instead, we used a ‘majority rule’ on the output of a panel of three expert human curators as a proxy for ground truth to train the model (Figure 2A). To maximize accuracy, human curators were not limited to viewing the cropped images. Instead, we performed whisker tracing and linking from full field images with the Janelia Whisker tracker \cite{clack_automated_2012}. We performed a first pass automated touch estimation based on extracted whisker curvature and estimated distance to pole. Humans then examined these estimated time series variables to screen and correct clear obvious misclassifications. Finally, for more challenging frames, humans inspected full, zoomed, and frame-by-frame difference images of touch and near touch periods with a visual browser and corrected scoring until satisfied. Human labels were applied to the test set of cropped images used for model evaluation. All three curators agreed on 99.46\% of test set frames.

Not all errors are created equal. We identified six distinct error types (Figure 2B). Four error types (splits, ghosts, misses, and joins) were classified as ‘touch-count errors’ because they affected the total number of touches. Touch count errors arise in four distinct ways. Misclassifying a frame in the middle of a touch event splits the touch into two separate events. Conversely, misclassifying a frame in the middle of a non-touch period creates a ghost touch, where no touch actually occurred. Failing to label any frames within a touch event results in a missed touch, while labeling a non-touch period between two touch events joins them into one. The other two errors are edge errors, where the length of a touch is shortened (deducts) or lengthened (appends) by mislabeling the start or end frames. These errors have different functional consequences for analysis of neural responses. Touch count errors propagate to touch evoked firing rates and peri-stimulus time histogram (PSTH) structure, while edge errors largely maintain touch evoked firing rate but degrade touch evoked latency and jitter measurements.

To determine human error rate we compared single curators against the consensus frames of the remaining two curators. Most human errors were edge errors, which occurred approximately 7 times more frequently than touch count errors (Figure 2C). During development of WhACC, model performance was evaluated across a range of error metrics, including area under the receiver operating characteristic curve (AUC), percent correct, touch count errors per touch (TC-error) and edge errors per touch. Our ultimate goal was to reduce TC-error because it most strongly degrades important measures like the total number of touches \cite{cheung_sensorimotor_2019}, angle at touch \cite{cheung_independent_2020}, and the number of spikes evoked on touch onset \cite{andrew_hires_low-noise_2015}.

\subsection{Selection of training, validation, and test set}
Our goal was to produce a reliable whisker-pole touch classifier with expert human level performance under a wide range of imaging conditions, all while minimizing the time burden of curating by hand. To support the generalization of model performance, our initial training and validation sets came from two experimentalists working on the same apparatus. In contrast, the test set comprised imaging data collected by two other people on different apparatus, across two labs and eight years. These datasets were used to train and select the base CNN model and the changes to the input images data (i.e. augmentation and lag images). For our final model, WhACC, we pooled all our data and generated new training, validation, and test sets to equally sample from the diversity of videos. To prevent overfitting, the data was assigned to each set based on video identity, ensuring that adjacent frames were included in only one dataset. To ensure reliable performance on new data, WhACC was validated on an additional holdout set of 16 different sessions of 100 trials each ($\sim$4 million frames), across different mice and different times, all of which were curated by a single curator. These data were not used for any stage of training, early stopping or feature selection and therefore serve as a reliable expectation of WhACC’s performance.

The training and validation sets were selected with a bias towards more touch-informative frames. In most imaged frames, the whisker is not present within the cropped window. These frames should be easily classified and are thought to provide mostly redundant training information. Thus, we underweighted their presence by removing many, but not all of them. To select more informative frames, we included all frames labeled as touch by human curation and any frames within 80 frames of a touch, discarding the rest (80-border extraction) (Figure 3A). This eliminated many, but not all ‘no-whisker frames’, while retaining all touch and most near-touch frames. Including some no-whisker frames was useful to provide baseline information for the negative class in the absence of a whisker. To increase model generalizability, we augmented 10 images for each original image. To maintain training and data storage efficiency, we used a stricter selection criterion for selecting these frames, which we restricted to all frames within three frames of any touch frame (Figure 3A). We hypothesized that these near-touch and touch frames would have much greater variation and therefore augmenting them would help the model generalize better. Overall the training set was $\sim$676,000 frames and validation set was $\sim$282,000 frames with a balanced set of classes (Figure 3B). The test dataset was not reduced at all and consisted of $\sim$780,000 frames that matched the distribution of a real dataset. In each dataset, only the frames where the object was within reach were used.

\begin{figure}[bt]
\centering
\includegraphics[width=\textwidth] {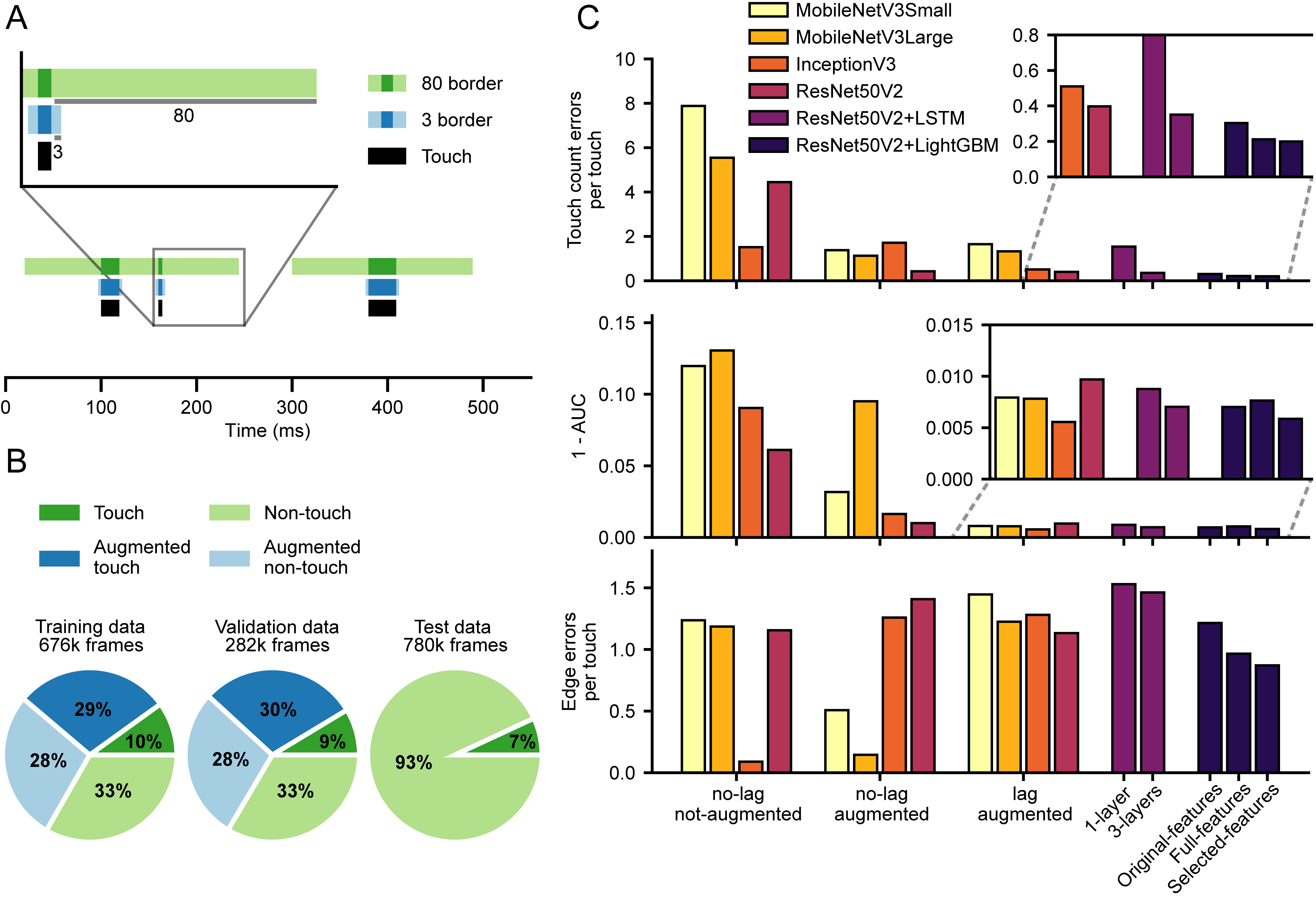}
\caption{\textbf{Data selection and model performance.}(A) Data selected for un-augmented (green) and augmented images (blue) using all frames within 80 or 3 frames from majority scored touch frames respectively. (B) Composition of training, validation and test datasets used to train and evaluate each model (C) Performance of four CNN models across three different image augmentation approaches, ResNet50V2 model features used as input into LSTMs, and ResNet50V2 model features used as input into LightGBM models before and after feature engineering and selection.}
\end{figure}

\begin{figure}[bt]
\centering
\includegraphics[width=\textwidth] {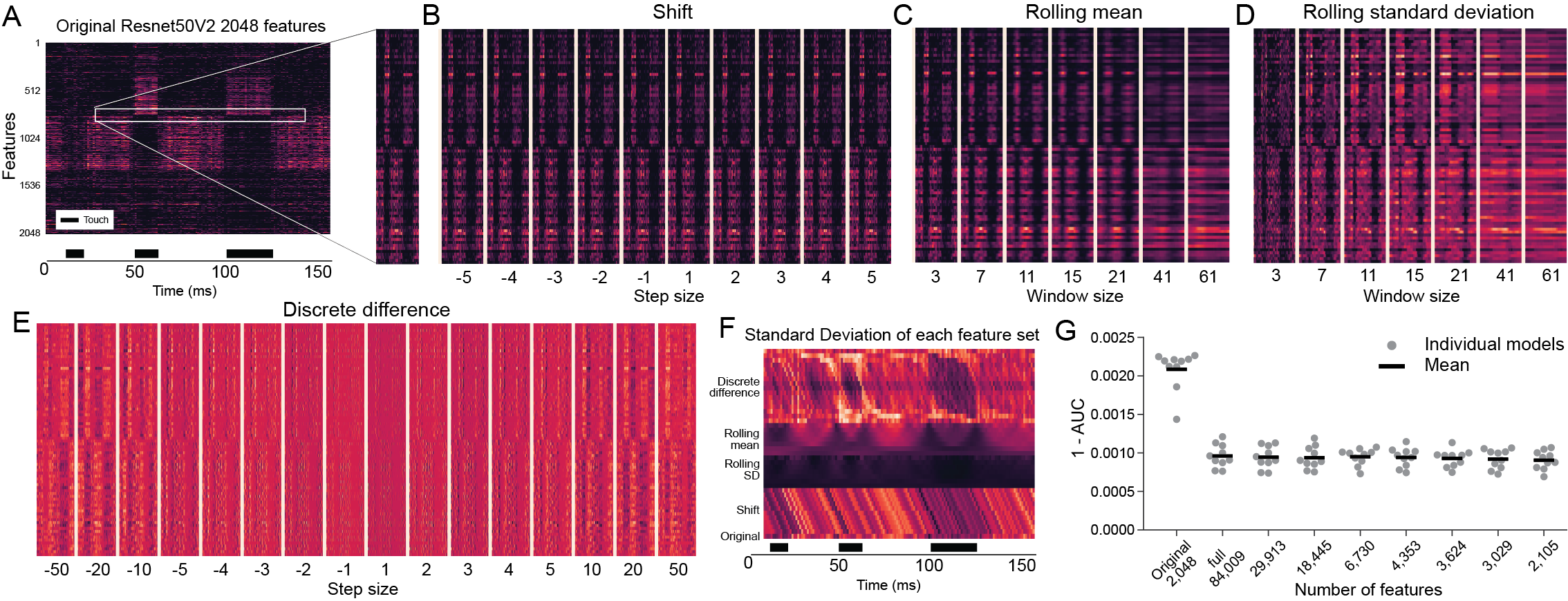}
\caption{\textbf{Feature engineering and selection.} A) The original 2,048 features extracted from the penultimate layer of Resnet 50 V2, (zoom) enlarged for detail (white box). Additional features generated by (B) shifting, (C) smoothing, (D) taking the rolling standard deviation, and (E) taking the discrete difference for each of the original 2,048 features. F) Standard deviation of the original and 40 additional engineered feature sets across feature space (columns). G) Model performance across feature engineering and reduction.}
\end{figure}

\begin{figure}[bt]
\centering
\includegraphics[width=\textwidth] {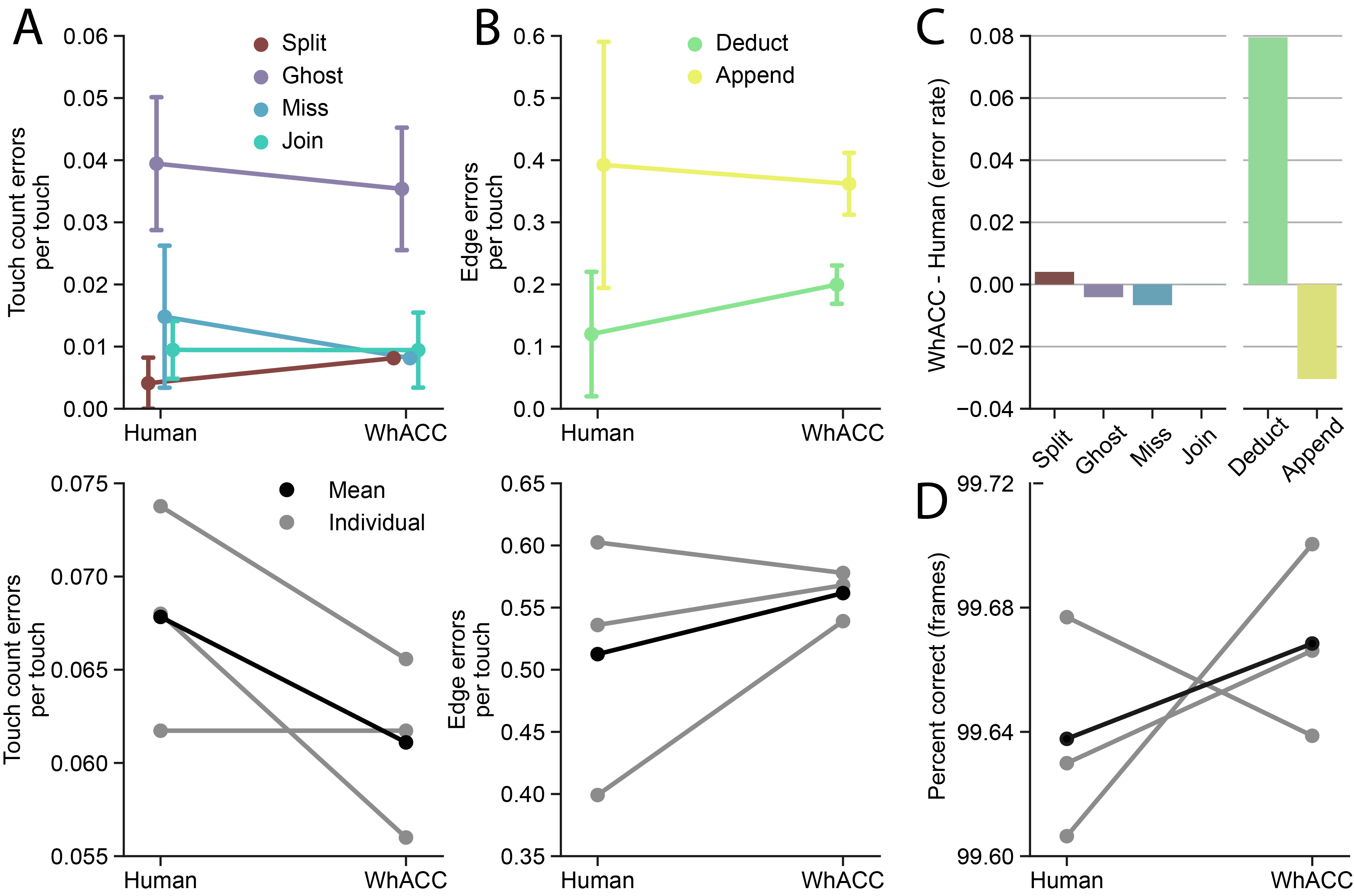}
\caption{\textbf{WhACC shows expert human level performance.} (A) Human vs WhACC touch count error rate for each error type (top) and in total (bottom), error bars indicate 95\% CI. (B) Same as A for edge errors. (C) Difference in error rate for human versus WhACC. Negative values indicate WhACC outperforming human curators on average. (D) Percent correct for individual and mean performance of human curators versus WhACC.}
\end{figure}

\begin{figure}[!ht]
\centering
\includegraphics[width=\textwidth] {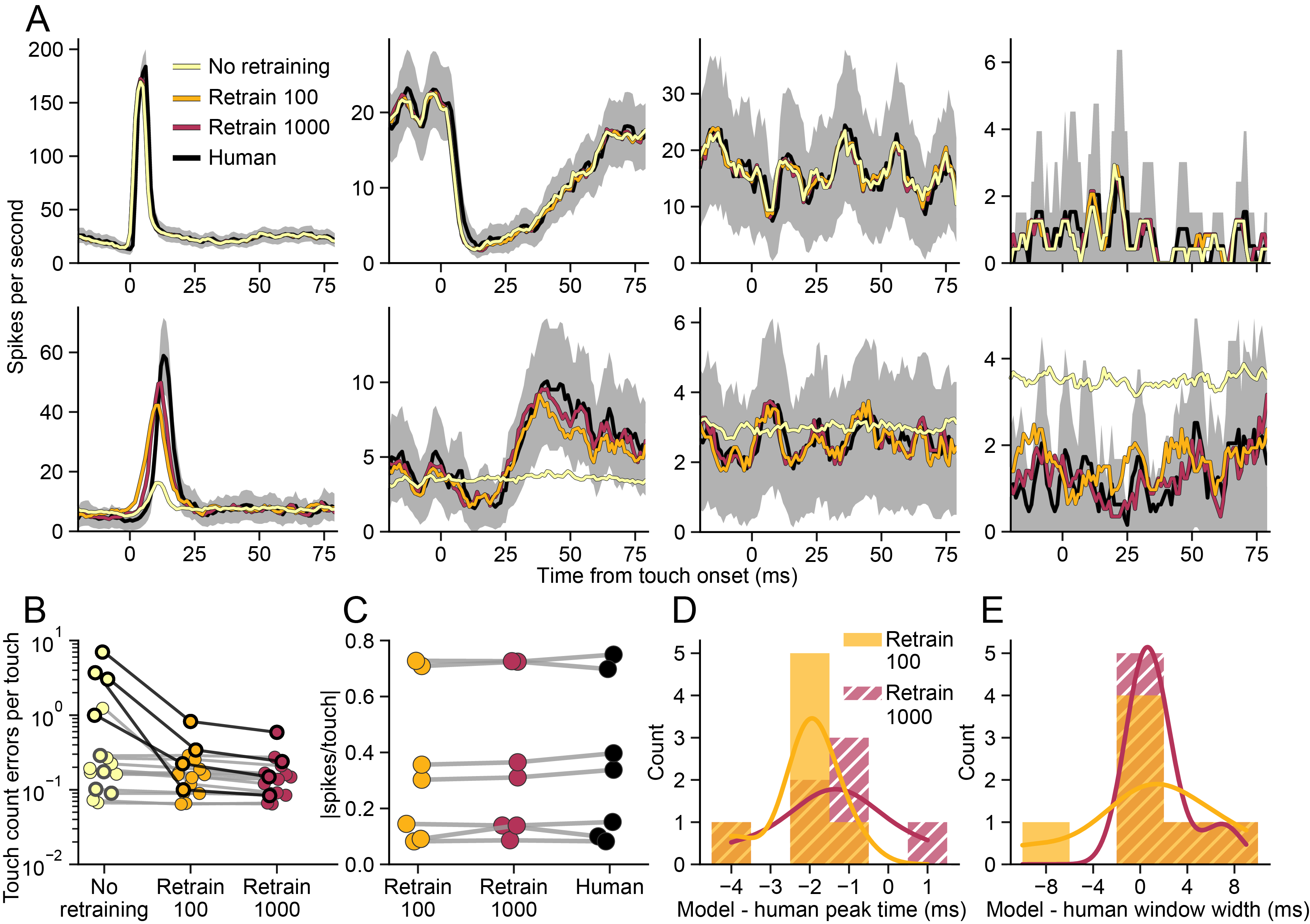}
\caption{\textbf{Retraining WhACC on a small sample of data can account for differences in datasets.} A) Touch aligned PSTHs curated by WhACC before retraining (yellow), after retraining on 100 frames (orange), and 1,000 frames (red) from each session compared against an expert curator (black, shaded regions represents 95\% CI). (Top row) Examples of high accuracy without retraining and (bottom row) high accuracy following retraining. PSTHs are ordered by ascending TC-errors before retraining. B) Touch count error rate from a holdout dataset consisting of 16 different sessions before and after retraining. Gray outlines indicate sessions from the top row in A and black outlines indicate those from the top row in A. C) Comparison of spike evoked touch count across human and retrained WhACC, n=7 touch responsive neurons. D) Difference in peak touch response times between retrained WhACC models and human curators E) Same as D for the width of touch response windows.}
\end{figure}

\begin{table}[!t] 
\centering 

\resizebox{\textwidth}{!}{ 
\begin{tabular}{ccccccccccc}
\rowcolor[HTML]{FFFFFF} 
\multicolumn{2}{c}{\cellcolor[HTML]{FFFFFF}\textbf{Images}} &
  \cellcolor[HTML]{FFFFFF} &
  \cellcolor[HTML]{FFFFFF} &
  \cellcolor[HTML]{FFFFFF} &
  \cellcolor[HTML]{FFFFFF} &
  \cellcolor[HTML]{FFFFFF} &
  \cellcolor[HTML]{FFFFFF} &
  \cellcolor[HTML]{FFFFFF} &
  \cellcolor[HTML]{FFFFFF} &
  \cellcolor[HTML]{FFFFFF} \\
\rowcolor[HTML]{FFFFFF} 
\textbf{Lag} &
  \textbf{Aug} &
  \multirow{-2}{*}{\cellcolor[HTML]{FFFFFF}\textbf{Base model name}} &
  \multirow{-2}{*}{\cellcolor[HTML]{FFFFFF}\textbf{TC-error}} &
  \multirow{-2}{*}{\cellcolor[HTML]{FFFFFF}\textbf{Edge errors per touch}} &
  \multirow{-2}{*}{\cellcolor[HTML]{FFFFFF}\textbf{AUC}} &
  \multirow{-2}{*}{\cellcolor[HTML]{FFFFFF}\textbf{Accuracy}} &
  \multirow{-2}{*}{\cellcolor[HTML]{FFFFFF}\textbf{Sensitivity}} &
  \multirow{-2}{*}{\cellcolor[HTML]{FFFFFF}\textbf{Specificity}} &
  \multirow{-2}{*}{\cellcolor[HTML]{FFFFFF}\textbf{Precision}} &
  \multirow{-2}{*}{\cellcolor[HTML]{FFFFFF}\textbf{Geometric Mean}} \\
\rowcolor[HTML]{FFFFFF} 
\cellcolor[HTML]{FFFFFF} &
  \cellcolor[HTML]{FFFFFF} &
  \textbf{MobileNetV3-Small} &
  7.901 &
  1.236 &
  0.762 &
  0.775 &
  0.485 &
  0.818 &
  0.282 &
  0.630 \\
\rowcolor[HTML]{FFFFFF} 
\cellcolor[HTML]{FFFFFF} &
  \cellcolor[HTML]{FFFFFF} &
  \textbf{MobileNetV3-Large} &
  5.558 &
  1.184 &
  0.741 &
  0.808 &
  0.413 &
  0.867 &
  0.314 &
  0.598 \\
\rowcolor[HTML]{FFFFFF} 
\cellcolor[HTML]{FFFFFF} &
  \cellcolor[HTML]{FFFFFF} &
  \textbf{Inception-v3} &
  \textbf{1.521} &
  \cellcolor[HTML]{FFF2CC}\textbf{0.088} &
  0.820 &
  \textbf{0.881} &
  0.130 &
  \textbf{0.992} &
  \textbf{0.696} &
  0.359 \\
\rowcolor[HTML]{FFFFFF} 
\multirow{-4}{*}{\cellcolor[HTML]{FFFFFF}\textbf{No}} &
  \multirow{-4}{*}{\cellcolor[HTML]{FFFFFF}\textbf{No}} &
  \textbf{ResNet50V2} &
  4.462 &
  1.156 &
  \textbf{0.879} &
  0.810 &
  \textbf{0.831} &
  0.807 &
  0.388 &
  \textbf{0.819} \\ \hline
\rowcolor[HTML]{FFFFFF} 
\cellcolor[HTML]{FFFFFF} &
  \cellcolor[HTML]{FFFFFF} &
  \textbf{MobileNetV3-Small} &
  2.214&
  \textbf{0.096} &
  0.623 &
  0.870 &
  0.125 &
  0.980 &
  0.478 &
  0.350 \\
\rowcolor[HTML]{FFFFFF} 
\cellcolor[HTML]{FFFFFF} &
  \cellcolor[HTML]{FFFFFF} &
  \textbf{MobileNetV3-Large} &
  5.094 &
  1.506 &
  0.896 &
  0.887 &
  0.763 &
  0.906 &
  0.544 &
  0.831 \\
\rowcolor[HTML]{FFFFFF} 
\cellcolor[HTML]{FFFFFF} &
  \cellcolor[HTML]{FFFFFF} &
  \textbf{Inception-v3} &
  1.757 &
  1.509 &
  \textbf{0.979} &
  0.945 &
  0.685 &
  \textbf{0.983} &
  \textbf{0.859} &
  0.821 \\
\rowcolor[HTML]{FFFFFF} 
\multirow{-4}{*}{\cellcolor[HTML]{FFFFFF}\textbf{Yes}} &
  \multirow{-4}{*}{\cellcolor[HTML]{FFFFFF}\textbf{No}} &
  \textbf{ResNet50V2} &
  \textbf{1.464} &
  1.465 &
  0.964 &
  \textbf{0.956} &
  \textbf{0.833} &
  0.975 &
  0.829 &
  \textbf{0.901} \\ \hline
\rowcolor[HTML]{FFFFFF} 
\cellcolor[HTML]{FFFFFF} &
  \cellcolor[HTML]{FFFFFF} &
  \textbf{MobileNetV3-Small} &
  1.387 &
  0.506 &
  0.937 &
  0.898 &
  0.242 &
  0.995 &
  0.870 &
  0.491 \\
\rowcolor[HTML]{FFFFFF} 
\cellcolor[HTML]{FFFFFF} &
  \cellcolor[HTML]{FFFFFF} &
  \textbf{MobileNetV3-Large} &
  1.135 &
  \textbf{0.142} &
  0.811 &
  0.875 &
  0.061 &
  \cellcolor[HTML]{FFF2CC}\textbf{0.995} &
  0.637 &
  0.246 \\
\rowcolor[HTML]{FFFFFF} 
\cellcolor[HTML]{FFFFFF} &
  \cellcolor[HTML]{FFFFFF} &
  \textbf{Inception-v3} &
  1.718 &
  1.259 &
  0.967 &
  0.922 &
  0.460 &
  0.990 &
  0.871 &
  0.675 \\
\rowcolor[HTML]{FFFFFF} 
\multirow{-4}{*}{\cellcolor[HTML]{FFFFFF}\textbf{No}} &
  \multirow{-4}{*}{\cellcolor[HTML]{FFFFFF}\textbf{Yes}} &
  \textbf{ResNet50V2} &
  \textbf{0.424} &
  1.409 &
  \textbf{0.980} &
  \textbf{0.965} &
  \textbf{0.848} &
  0.982 &
  \textbf{0.877} &
  \textbf{0.913} \\ \hline
\cellcolor[HTML]{FFFFFF} &
  \cellcolor[HTML]{FFFFFF} &
  \cellcolor[HTML]{FFFFFF}\textbf{MobileNetV3-Small} &
  \cellcolor[HTML]{FFFFFF}1.645 &
  \cellcolor[HTML]{FFFFFF}1.446 &
  \cellcolor[HTML]{FFFFFF}0.984&
  \cellcolor[HTML]{FFFFFF}0.956 &
  \cellcolor[HTML]{FFFFFF} \textbf{0.959} &
  \cellcolor[HTML]{FFFFFF}0.956 &
  \cellcolor[HTML]{FFFFFF}0.761 &
  \cellcolor[HTML]{FFFFFF}0.957 \\
\rowcolor[HTML]{FFFFFF} 
\cellcolor[HTML]{FFFFFF} &
  \cellcolor[HTML]{FFFFFF} &
  \textbf{MobileNetV3-Large} &
  1.326 &
  1.225 &
  0.984 &
  0.963 &
  0.893 &
  0.973 &
  0.829 &
  0.932 \\
\rowcolor[HTML]{FFFFFF} 
\cellcolor[HTML]{FFFFFF} &
  \cellcolor[HTML]{FFFFFF} &
  \textbf{Inception-v3} &
  0.512 &
  1.281 &
  \cellcolor[HTML]{FFF2CC} \textbf{0.989} &
  \textbf{0.981} &
  0.930 &
  \textbf{0.988} &
  \cellcolor[HTML]{FFF2CC} \textbf{0.920} &
  \textbf{0.959} \\

\rowcolor[HTML]{FFFFFF} 
\cellcolor[HTML]{FFFFFF} &
  \cellcolor[HTML]{FFFFFF} &
  \textbf{ResNet50V2} &
  \textbf{0.402} &
  \textbf{1.131} &
  0.981 &
  0.977 &
  0.930 &
  0.984 &
  0.894 &
  0.957 \\ \hline
\rowcolor[HTML]{FFFFFF} 
\multirow{-5}{*}{\cellcolor[HTML]{FFFFFF}\textbf{Yes}} &
 \multirow{-5}{*}{\cellcolor[HTML]{FFFFFF}\textbf{Yes}} &
  \textbf{ResNet50V2 + Single-layer LSTM} &
  1.545 &
  1.529 &
  0.983 &
  0.905 &
  \cellcolor[HTML]{FFF2CC}\textbf{0.963} &
  0.897 &
  0.579 &
  0.929 \\
  \rowcolor[HTML]{FFFFFF} 
\cellcolor[HTML]{FFFFFF} &
  \cellcolor[HTML]{FFFFFF} &
  \textbf{ResNet50V2 + Multi-layer LSTM} &
  0.354 &
  1.462 &
  0.986 &
  0.970 &
  0.957 &
  0.972 &
  0.834 &
  0.964 \\
  \rowcolor[HTML]{FFFFFF} 
\cellcolor[HTML]{FFFFFF} &
  \cellcolor[HTML]{FFFFFF} &
  \textbf{ResNet50V2 + LightGBM Original (2,048)} &
  0.307 &
  1.214 &
  0.986 &
  0.978 &
  0.927 &
  \textbf{0.986} &
  0.906 &
  0.956 \\
  \rowcolor[HTML]{FFFFFF} 
\cellcolor[HTML]{FFFFFF} &
  \cellcolor[HTML]{FFFFFF} &
  \textbf{ResNet50V2 + LightGBM Full (84,006)} &
  0.214 &
  \textbf{0.966} &
  0.985 &
  0.980 &
  0.942 &
  \textbf{0.986} &
  0.907 &
  0.964 \\
  \rowcolor[HTML]{FFFFFF} 
\multirow{-5}{*}{\cellcolor[HTML]{FFFFFF}\textbf{Yes}} &
 \multirow{-5}{*}{\cellcolor[HTML]{FFFFFF}\textbf{Yes + No}} &
  \textbf{ResNet50V2 + LightGBM selected (2,105)} &
   \cellcolor[HTML]{FFF2CC}\textbf{0.202} &
  \ 1.870 &
  \textbf{0.988} &
   \cellcolor[HTML]{FFF2CC}\textbf{0.982} &
  0.954 &
  \textbf{0.986} &
  \textbf{0.911} &
   \cellcolor[HTML]{FFF2CC}\textbf{0.970} \\
\end{tabular}
} 
 \normalsize

    \caption{\textbf{Performance metrics across models after median smoothing.} Performance metrics across all tested model variants. Bold text indicates the best performance for that metric across each image modification approach, yellow highlighted text indicates the best performance for that metric across all model variants. ‘Lag’ and ‘Aug’ indicate that lag and augmented images were included in training these models, respectively. All metrics were calculated after median smoothing predictions with a window of five. These results are from the test set outlined in Figure 3B, that consists of visually different data from another laboratory. Parentheses indicate the number of features from each frame inputted into the LightGBM model.}
    \label{tab:my_label}
\end{table}

\begin{table}[!t] 

\centering 
\resizebox{\textwidth}{!}{ 
\begin{tabular}{ccccccccccc}
\rowcolor[HTML]{FFFFFF} 
\multicolumn{2}{c}{\cellcolor[HTML]{FFFFFF}\textbf{Images}} &
  \cellcolor[HTML]{FFFFFF} &
  \cellcolor[HTML]{FFFFFF} &
  \cellcolor[HTML]{FFFFFF} &
  \cellcolor[HTML]{FFFFFF} &
  \cellcolor[HTML]{FFFFFF} &
  \cellcolor[HTML]{FFFFFF} &
  \cellcolor[HTML]{FFFFFF} &
  \cellcolor[HTML]{FFFFFF} &
  \cellcolor[HTML]{FFFFFF} \\
\rowcolor[HTML]{FFFFFF} 
\textbf{Lag} &
  \textbf{Aug} &
  \multirow{-2}{*}{\cellcolor[HTML]{FFFFFF}\textbf{Base model name}} &
  \multirow{-2}{*}{\cellcolor[HTML]{FFFFFF}\textbf{TC-error}} &
  \multirow{-2}{*}{\cellcolor[HTML]{FFFFFF}\textbf{Edge errors per touch}} &
  \multirow{-2}{*}{\cellcolor[HTML]{FFFFFF}\textbf{AUC}} &
  \multirow{-2}{*}{\cellcolor[HTML]{FFFFFF}\textbf{Accuracy}} &
  \multirow{-2}{*}{\cellcolor[HTML]{FFFFFF}\textbf{Sensitivity}} &
  \multirow{-2}{*}{\cellcolor[HTML]{FFFFFF}\textbf{Specificity}} &
  \multirow{-2}{*}{\cellcolor[HTML]{FFFFFF}\textbf{Precision}} &
  \multirow{-2}{*}{\cellcolor[HTML]{FFFFFF}\textbf{Geometric Mean}} \\
\rowcolor[HTML]{FFFFFF} 
\cellcolor[HTML]{FFFFFF} &
  \cellcolor[HTML]{FFFFFF} &
  \textbf{MobileNetV3-Small} &
  25.907 &
  1.447 &
  0.747 &
  0.755 &
  0.497 &
  0.793 &
  0.262 &
  0.628 \\
\rowcolor[HTML]{FFFFFF} 
\cellcolor[HTML]{FFFFFF} &
  \cellcolor[HTML]{FFFFFF} &
  \textbf{MobileNetV3-Large} &
  17.153 &
  1.190 &
  0.727 &
  0.793 &
  0.421 &
  0.848 &
  0.290 &
  0.598 \\
\rowcolor[HTML]{FFFFFF} 
\cellcolor[HTML]{FFFFFF} &
  \cellcolor[HTML]{FFFFFF} &
  \textbf{Inception-v3} &
  \textbf{3.238} &
  \textbf{0.280} &
  0.791 &
  \textbf{0.880} &
  0.126 &
  \textbf{0.991} &
  \textbf{0.667} &
  0.354 \\
\rowcolor[HTML]{FFFFFF} 
\multirow{-4}{*}{\cellcolor[HTML]{FFFFFF}\textbf{No}} &
  \multirow{-4}{*}{\cellcolor[HTML]{FFFFFF}\textbf{No}} &
  \textbf{ResNet50V2} &
  18.941 &
  1.182 &
  \textbf{0.866} &
  0.792 &
  \textbf{0.813} &
  0.789 &
  0.362 &
  \textbf{0.801} \\ \hline
\rowcolor[HTML]{FFFFFF} 
\cellcolor[HTML]{FFFFFF} &
  \cellcolor[HTML]{FFFFFF} &
  \textbf{MobileNetV3-Small} &
  8.292 &
  \cellcolor[HTML]{FFF2CC}\textbf{0.182} &
  0.618 &
  0.857 &
  0.128 &
  0.964 &
  0.343 &
  0.351 \\
\rowcolor[HTML]{FFFFFF} 
\cellcolor[HTML]{FFFFFF} &
  \cellcolor[HTML]{FFFFFF} &
  \textbf{MobileNetV3-Large} &
  21.908 &
  1.463 &
  0.868 &
  0.848 &
  0.741 &
  0.864 &
  0.445 &
  0.800 \\
\rowcolor[HTML]{FFFFFF} 
\cellcolor[HTML]{FFFFFF} &
  \cellcolor[HTML]{FFFFFF} &
  \textbf{Inception-v3} &
  8.860 &
  1.598 &
  \textbf{0.957} &
  0.932 &
  0.661 &
  \textbf{0.972} &
  \textbf{0.779} &
  0.802 \\
\rowcolor[HTML]{FFFFFF} 
\multirow{-4}{*}{\cellcolor[HTML]{FFFFFF}\textbf{Yes}} &
  \multirow{-4}{*}{\cellcolor[HTML]{FFFFFF}\textbf{No}} &
  \textbf{ResNet50V2} &
  \textbf{7.992} &
  1.371 &
  0.950 &
  \textbf{0.941} &
  \textbf{0.809} &
  0.960 &
  0.749 &
  \textbf{0.881} \\ \hline
\rowcolor[HTML]{FFFFFF} 
\cellcolor[HTML]{FFFFFF} &
  \cellcolor[HTML]{FFFFFF} &
  \textbf{MobileNetV3-Small} &
  3.565 &
  0.884 &
  0.921 &
  0.897 &
  0.255 &
  0.992 &
  0.826 &
  0.503 \\
\rowcolor[HTML]{FFFFFF} 
\cellcolor[HTML]{FFFFFF} &
  \cellcolor[HTML]{FFFFFF} &
  \textbf{MobileNetV3-Large} &
  \textbf{1.743} &
  \textbf{0.239} &
  0.803 &
  0.875 &
  0.064 &
  \cellcolor[HTML]{FFF2CC}\textbf{0.995} &
  0.638 &
  0.253 \\
\rowcolor[HTML]{FFFFFF} 
\cellcolor[HTML]{FFFFFF} &
  \cellcolor[HTML]{FFFFFF} &
  \textbf{Inception-v3} &
  6.504 &
  1.197 &
  0.947 &
  0.921 &
  0.478 &
  0.986 &
  0.836 &
  0.686 \\
\rowcolor[HTML]{FFFFFF} 
\multirow{-4}{*}{\cellcolor[HTML]{FFFFFF}\textbf{No}} &
  \multirow{-4}{*}{\cellcolor[HTML]{FFFFFF}\textbf{Yes}} &
  \textbf{ResNet50V2} &
  1.987 &
  1.468 &
  \textbf{0.976} &
  \textbf{0.962} &
  \textbf{0.848} &
  0.979 &
  \textbf{0.858} &
  \textbf{0.911} \\ \hline
\cellcolor[HTML]{FFFFFF} &
  \cellcolor[HTML]{FFFFFF} &
  \cellcolor[HTML]{FFFFFF}\textbf{MobileNetV3-Small} &
  \cellcolor[HTML]{FFFFFF}7.585 &
  \cellcolor[HTML]{FFFFFF}1.321 &
  \cellcolor[HTML]{FFFFFF}0.979 &
  \cellcolor[HTML]{FFFFFF}0.943 &
  \cellcolor[HTML]{FFFFFF} \textbf{0.948} &
  \cellcolor[HTML]{FFFFFF}0.942 &
  \cellcolor[HTML]{FFFFFF}0.707 &
  \cellcolor[HTML]{FFFFFF}0.945 \\
\rowcolor[HTML]{FFFFFF} 
\cellcolor[HTML]{FFFFFF} &
  \cellcolor[HTML]{FFFFFF} &
  \textbf{MobileNetV3-Large} &
  7.226 &
  \textbf{1.089} &
  0.975 &
  0.947 &
  0.870 &
  0.959 &
  0.757 &
  0.913 \\
\rowcolor[HTML]{FFFFFF} 
\cellcolor[HTML]{FFFFFF} &
  \cellcolor[HTML]{FFFFFF} &
  \textbf{Inception-v3} &
  3.254 &
  1.261 &
  \textbf{0.983} &
  \textbf{0.974} &
  0.905 &
  \textbf{0.984} &
  \textbf{0.895} &
  0.944 \\
\rowcolor[HTML]{FFFFFF} 
\multirow{-4}{*}{\cellcolor[HTML]{FFFFFF}\textbf{Yes}} &
  \multirow{-4}{*}{\cellcolor[HTML]{FFFFFF}\textbf{Yes}} &
  \textbf{ResNet50V2} &
  \textbf{2.575} &
  1.166 &
  0.977 &
  0.971 &
  0.925 &
  0.978 &
  0.860 &
  \textbf{0.951} \\ \hline
\rowcolor[HTML]{FFFFFF} 
\cellcolor[HTML]{FFFFFF} &
  \cellcolor[HTML]{FFFFFF} &
  \textbf{ResNet50V2 +   Single-layer LSTM} &
  2.546 &
  1.572 &
  0.982 &
  0.904 &
  \cellcolor[HTML]{FFF2CC}\textbf{0.963} &
  0.895 &
  0.576 &
  0.929 \\ 
\rowcolor[HTML]{FFFFFF} 
\cellcolor[HTML]{FFFFFF} &
  \cellcolor[HTML]{FFFFFF} &
  \textbf{ResNet50V2   + Muli-layer LSTM} &
  \cellcolor[HTML]{FFF2CC}\textbf{0.411} &
  1.478 &
  0.986 &
  0.970 &
  0.957 &
  0.972 &
  0.833 &
  0.964 \\
\rowcolor[HTML]{FFFFFF} 
\cellcolor[HTML]{FFFFFF} &
  \cellcolor[HTML]{FFFFFF} &
  \textbf{ResNet50V2   + LightGBM Original (2,048)} &
  1.513 &
  1.229 &
  0.983 &
  0.976 &
  0.919 &
  0.984 &
  0.896 &
  0.951 \\ 
\rowcolor[HTML]{FFFFFF} 
\cellcolor[HTML]{FFFFFF} &
  \cellcolor[HTML]{FFFFFF} &
  \textbf{ResNet50V2   + LightGBM Full (84,006)} &
  0.478 &
  0.994 &
  0.984 &
  0.980 &
  0.942 &
  0.985 &
  0.905 &
  0.963 \\
\rowcolor[HTML]{FFFFFF} 
\multirow{-5}{*}{\cellcolor[HTML]{FFFFFF}\textbf{Yes}} &
  \multirow{-5}{*}{\cellcolor[HTML]{FFFFFF}\textbf{ Yes + No}} &
  \textbf{ResNet50V2   + LightGBM selected (2,105)} &
  0.483 &
  \textbf{0.888} &
  \cellcolor[HTML]{FFF2CC}\textbf{0.988} &
  \cellcolor[HTML]{FFF2CC}\textbf{0.982} &
  0.954 &
  \textbf{0.986} &
  \cellcolor[HTML]{FFF2CC}\textbf{0.908} &
  \cellcolor[HTML]{FFF2CC}\textbf{0.970} \\ 
\end{tabular} 
} 
 \normalsize
    \caption{\textbf{Performance metrics across models without median smoothing.} Same as Table 1 but without median smoothing.}
\end{table}

\subsection{Model selection and evaluation}
To assess the feasibility of automated contact curation, we trained and tested four different fully unfrozen base models: ResNet50V2 \cite{He_IdentityMapping_2016}, Inception-v3 \cite{szegedy_rethinking_2015}, MobileNetV3-Large, and MobileNetV3-Small \cite{howard_searching_2019}. We selected these four models because they are relatively lightweight, are pre-trained on ImageNet, and have a well-established proven track record in addressing a wide range of problems across various domains. For each base model we either used the original images or a set of augmented images combined with the original images. Finally, for each combination of these, we compared models trained with original images versus images where the three-color channels contained frames from times t, t-1 and t-2 (Figure 1B, left). We refer to these as lag images because they contain frames that lag behind the timepoint we are predicting. Out of the resulting 16 unique models evaluated in Table 1 and 2, we discuss 12 in detail, focusing on the ones that showed the greatest improvement with each sequential step. The remaining four models trained on lag images with no augmentation were excluded from the discussion as they did not perform as well as augmented images alone, but still showed improvement over original images. TC-error and AUC were moderately correlated (r = -0.52), consequently there were cases where the model with the highest AUC was not the same as the model with the lowest TC-error. Due to WhACC’s intended use case, all models were selected based solely on TC-error to emphasize temporal consistency. 

Training each model using single time point images without augmentation showed poor results overall. The best model, Inception-v3 achieving a TC-error of 1.521 and an AUC of 0.820 (Table 1). Adding augmented images to the training data decreased TC-error in all base models except Inception-v3. ResNet50V2 performed best for this variation, with a TC-error 0.424 and an AUC of 0.980 (Table 1). Finally using the same set of augmented and original images, we changed the input images to contain frames from previous two time points in the color channels to predict timepoint corresponding to the leading frame. Incorporating this temporal information decreased TC-error for the deeper models, Inception-v3 and ResNet50V2. Interestingly, for both MobileNetV3 models, adding temporal information via lag images decreased performance measured by TC-error, but increased performance measured by AUC. The increased TC-error could be due to the depthwise separable convolution architecture that makes MobileNet so efficient. The best performing model again was ResNet50V2, with a TC-error of 0.402 and corresponding AUC of 0.981. Despite having a better AUC of 0.989, Inception-v3 had a slightly worse TC-error of 0.512 (Table 1). Therefore we selected ResNet50V2 as the best model for our purposes.

While TC-error is the most important measure, edge errors per touch can also negatively impact analysis of neural data. Edge errors per touch were slightly negatively correlated with TC-error (r = -0.11). Examining image modification approaches revealed that on average, augmentation with lag images showed the best performance for TC-error, but the worst performance for edge count errors; this suggests that there is some tradeoff between these two types of errors (Table 1).

The above results showed that our overall approach to automated touch classification was feasible but needed improvements to match expert human level performance. To improve the overall performance, we implemented a two-stage hybrid model. The first stage of the model involved extracting 2,048 features from the penultimate layer of the best performing model, which was ResNet50V2 trained on augmentation and lag images. These features were then inputted into a LightGBM classifier \cite{ke_lightgbm:_2017}. We chose to use LightGBM due to its excellent performance across a range of classification and regression problems, as well as its fast and lightweight nature. Preliminary results showed a noticeable increase in overall performance, reducing TC-errors to 0.307 (Table 1, Figure 3C), however, to achieve expert level performance, we sought insights from human curators to gain further understanding of how to improve the system based on temporal information.

So far, our approach to touch classification was limited to integrating temporal information over three frames. However, expert human curators integrate information over longer periods of time. A curator can intuitively infer times of high touch probability to bias their choice, since touches occur in clusters. They can also bias their confidence of a touch frame based on previously identified touch frames, since touches most often occur in segments of many frames in a row. For example, if the last 20 frames were identified as touch frames, the likelihood of the next frame being a touch frame is much higher than a randomly sampled frame. Moreover, curators can distinguish onset and offset touch frames by identifying sudden changes in features like whisker speed and whisker bending. They do this by comparing them to earlier frames that act as a visual baseline for this change. Currently, our model does not have access to these features, so we engineered features to capture more of them.

Building off the original 2,048 extracting features, we engineered features which improved model performance (Figure 3C). How we engineered features was informed by the strategies of expert curators outlined above. They included forward and back shifts of up to 5 frames, rolling means and rolling standard deviation of windows ranging from 3 to 61 frames, and discrete differences of frames ranging from -50 to 50 frames apart (Methods). Frame shifting and small window rolling mean operations should inform the current frame based on features of surrounding frames (Figure 4B, 4C). Larger window rolling mean and larger window rolling standard deviation should contain information of regions of high touch probability or clusters of touches (Figure 4C, 4D). Smaller window rolling standard deviation and discrete difference can reveal sudden changes in features which can help identify onset and offset times (Figure 4D, 4E). Finally, by taking the standard deviation across feature space, we get a measure of dispersal unique to each feature set, each of which can inform touch properties in their own way (Figure 4F). 

Combined with the original features we now had a total of 84,009 features. To reduce model complexity and memory demands, we reduced the total number of features using recursive feature elimination (Methods). Our original training data could not fit into memory with all the features, so we divided data into 10 training and validation sets and fit 10 light GBM classifier models for each feature selection step. To ensure we could capture as many high value features as possible, we split our data across all sessions, therefore these performance metrics are not directly comparable to previous figures (Methods). First, each model was trained with the original 2,048 features. Next we trained each model on the full set of 84,009 features. Adding engineered features reduced error by over 50\% (Figure 4G). For the full feature set model, most features had embedded feature importance of 0 for all 10 models (i.e., not a single model used these features), so we eliminated all these features for the next iteration, leaving 28,913 remaining. We continued to reduce features by selecting features that passed a threshold until there was a negative impact on mean performance of the 10 models. In total, we isolated 2,105 high value features. The mean performance of the models trained on this selection was indistinguishable from those trained on the full feature set (Figure 4G).

To compare our approach to other model architectures, we trained a single layer long short-term memory (LSTM) and a multi-layer LSTM on the 2,048 ResNet50V2 features (Methods). We compared these models to LightGBM models trained on the original 2,048 features before feature engineering (original-feature model), the full 84,009 features after feature engineering (full-feature model), and the 2,105 high-value features after feature selection (selected-feature model; Figure 3C). Note that some test data was used for feature selection, thus the selected-feature model has some data leakage by way of which features were included. There is no leakage in the full-feature model. Multiple attempts were made using single layer LSTM (Figure 3C) and gated recurrent unit (not shown) models to capture temporal information, but all attempts proved unsuccessful at translating to the test dataset and had higher TC-error than the ResNet50V2 model alone. One potential explanation for this is the visual differences between the training/validation data and the test data. Further attempts at a more complex multi-layer LSTM with dropout layers during training showed a marked increase in performance (Methods). Prior to median smoothing this model even showed slightly lower TC-errors compared to the full-feature model, but more edge errors (Table 2). However, after median smoothing, even the original-feature model, with no additional temporal information performed better than the multi-layer LSTM. This suggests that the increased performance from the multi-layer LSTM was largely due to reducing TC-errors of length 1 to 2 frames long, which can easily be corrected for using median smoothing window of five.

Compared to the multi-layer LSTM with TC-error of 0.35, the original-feature LightGBM model showed a 13\% reduction, the full-feature model showed a 39\% reduction, and the selected-feature model showed a 43\% reduction in TC-errors. Among these methods the multi-layer LSTM also had to most edge-errors per touch. Taken together LightGBM has a clear advantage in both performance but also in the speed and ease in retraining for future datasets. 

Using the selected 2,105 features, we trained a final LightGBM classifier using Optuna \cite{akiba_optuna:_2019}, to optimize the hyperparameters over 100 Optuna trials (Methods). The final model, which we named WhACC, was trained on a dataset consisting of data split from all sessions and is therefore different from the selected-feature model described above. To compare WhACC with expert curators, we compared all consensus frames of two curators against either the other curator or against WhACC. We did this for each combination of two curators for a total of three comparisons. This method of comparison reveals the human error rate and establishes a realistic performance ceiling. Overall, WhACC made fewer TC-errors on average. For each type of touch count errors, WhACC either had similar or less variability compared to the human error rates (Figure 5A). WhACC made more edge errors per touch on average, specifically because of more deduct errors. For both types of edge errors WhACC showed less variability compared to expert curators (Figure 5B). On average WhACC performs as well as a human curator or slightly better when evaluating on TC-errors but has a slight bias towards shortening the length of touches (Figure 5C). Finally, WhACC shows a nearly identical but slightly larger mean percent correct frames compared to expert curators (Figure 5D). 

WhACC performs comparably to an expert curator on our current data sets. However, given the variability of video data between labs, experimental apparatuses, experimenters, and over time, we need to ensure its effectiveness on future datasets (e.g., data drift). To account for these differences, we developed a retraining system. First, we automatically sample video frames based on user-defined time when the pole is within reach. Using the tracked object position data, frames are automatically sampled equally based on the object's location in each video. Next, we curate this small subset using the included GUI. Finally, we retrain the LightGBM model with the newly curated data combined with the original training data. This approach enables us to quickly adapt WhACC to new datasets, without the need to retrain the ResNet50V2 model or repeat the time-intensive feature selection step. We then evaluated our retraining procedure using a holdout dataset with 16 different recording sessions of 100 videos each for a total of $\sim$ 4 million frames.

We validated our retraining procedure by comparing WhACC before retraining, after retraining with 100 frames per session (.03\% of the data), and after retraining with 1,000 frames per session (.3\% of the data). Curating 1,000 frames takes between two and four minutes, while curating an entire session would take between three and five hours of focused work. Before retraining, WhACC performed well on 11 sessions (Figure 6A, top row) but poorly on the other 5 (Figure 6A, bottom row). Retraining with 100 frames per session led to a large reduction in TC-errors for the 5 poor performing sessions, while the 1000-frame model yielded an additional incremental reduction for some high-error sessions (Figure 6B). Both WhACC retrained models and the human curator identified the same seven sessions corresponding to touch-responsive neurons. The 1000-frame model generated more spikes per touch for three of the seven touch neurons compared to human curated data (Figure 6C). WhACC peak responses tended to be 1 to 2 ms earlier than the human curator's due to the increased number of deduct errors (Figure 6D). Finally, we found the signal window size of the 1000-frame model matched that of the human curator for five neurons but was slightly larger for the other two (Figure 6E). Taken together, we have shown the utility of WhACC as a high-speed video touch classification system that allows for reliable, consistent, and accurate predictions that can be updated to adapt to new datasets.

\section{Discussion}
\subsection{Summary of WhACC}
The present study describes a novel approach to classify whisker touch frames from high-speed video using a 2-stage hybrid model implemented in a Python package named WhACC. WhACC was trained based on majority labels established by three expert curators and evaluated on a custom measure which penalized error classes which would be most detrimental to our downstream electrophysiology analysis. We integrated some temporal information into ResNet50V2 by layering different frames in color channels and trained the model using augmented images and dropout to increase generalizability. To further optimize performance, we extracted, engineered, and selected features from ResNet50V2. These features were fed into a LightGBM classifier, and the hyperparameters were optimized through OPTUNA. To account for data drift (e.g., changing imaging conditions over time), we developed and empirically validated a retraining system using a small number of sample frames. Our findings demonstrate that WhACC is an effective and reliable tool for whisker touch classification that can save time compared to manual curation.

\subsection{Potential limitations}
WhACC achieves human expert-level performance on our data but there are several conditions where its performance is uncertain. First, because the ResNet50V2 model was trained on frames from 1,000 frames per second (fps) video, WhACC may not be as effective for other frame rates because the lag images are constructed from differently sized time-steps. Second, WhACC may struggle when multiple whiskers are in the frame simultaneously, as it was trained to classify touches from single whisker videos. Third, our model was only trained on images where the object was a small round pole so how well it generalizes to other contact objects is not known. For videos containing other objects, possible touch points (around the object) must still be contained within the extraction window (between 61x61 and 96x96 pixels). For example, an object spanning the entire viewing area is incompatible with WhACC without major modification.  Finally, we must consider video clarity and contrast because this impacts human curation, whisker tracing in the Janelia whisker tracker, and WhACC. 

\subsection{Videos with different frame rates}
Preliminary tests using low fps video in our lab show promise for accurately identifying touches. Even so, an option for improvement could be to modify lower fps videos using a frame interpolation model to generate intermediate frames to match our training data at 1,000 fps. On the contrary, generating features from a low fps video and then interpolating those features afterward may not work as intended, because some of the features likely depend on the frame rate of the lag images (e.g., a velocity related feature derived from the distance of the whiskers in different color channels). Another alternative could be to downsample the original training data and retrain WhACC from scratch. 

\subsection{Multi-whisker video and alternative contact objects}
Despite training on single-whisker videos, WhACC's ResNet50V2 stage likely captures general features useful for predicting touches in multi-whisker video. We believe that the existing retraining procedure should work well, provided that the training data contains images with multi-whisker touches. For different contact objects, our frame extraction method should work without issue because it is based on a template image defined by the user. While it's unclear how well WhACC will perform on contact objects of different shapes or sizes post-retraining, we believe it should generalize well for at least a subset of possible objects, given our use of augmented training images and poles of varying diameters and optical zooms. Finally, we expect WhACC to perform well in freely moving rodents touching static objects from different directions, because we trained on rotated images, although we have not tested this.

\subsection{Why is retraining required?}
We determined that the reason WhACC failed on five sessions prior to retraining was due to very poor performance on whisker out-of-frame images, which was unexpected given that these frames should be easy to classify. However, as soon as the whisker came into view, WhACC performed well. This result could be attributed to two main factors. First, it could be due to the variability of what ‘baseline’ whisker out-of-frame images look like across sessions, including changes in focus, contrast and even debris on the pole and whisker. Second, relatively few whisker out-of-frame images were included in our augmented training data, which most likely hindered WhACC’s ability to generalize to these frames. In addition, we noticed that WhACC performed slightly worse on far distance touches, when only the tip of the whisker touched the pole. We suspect this is due to limited training data for these scenarios and because of a low signal to noise of the visibility of the thin tapered end of the whisker.

\subsection{Final considerations and related work}
We use multiple approaches to reduce TC-errors including data augmentation, model selection, feature engineering, model stacking, and median smoothing predictions. Median smoothing predictions consistently decreases TC-error across all models we tested. This is consistent with how TC-errors are calculated. For example, touch segments are commonly between 10 and 200 frames long and even a single misclassified frame within these touch segments will lead to a touch count error, despite all the other surround frames being correct. Smoothing is helpful only in these scenarios where infrequent and isolated errors occur, but not in cases where errors occur over longer timescales. Infrequent and isolated errors often occur during an onset or offset event where the transition from one class to the other can be noisy, resulting in touch count errors of one or two frames. This helps to explain the large impact median smoothing has on reducing TC-error. 

Feature engineering also accounts for large reduction in TC-error. Consider the original-feature model with 2,408 features in table 1 (TC-error 1.513). Applying only smoothing to this model decreases TC-errors by 80\% (TC-error 0.307) compared to using feature engineering without smoothing which decreases TC-errors by 68\% (TC-error 0.483). Importantly, using both smoothing and feature engineering the TC-error is reduced to 0.202 indicating that the combination of these methods provides complimentary information. 

TC-error measures temporal consistency of predictions by penalizing changes in total touch count, rather than only misclassified individual frames. After training with lag and augmentation images and median smoothing, we show that TC-errors decreased for Inception-v3 and ResNet50V2 compared to models trained on only augmented images. The reverse was true for both MobileNet models. This might be related to their depthwise separable convolution architecture which first uses depthwise convolution on each color channel separately and then uses pointwise convolution on the results from the depthwise convolution \cite{chollet_xception_2017}. This method is an efficient way to independently map spatial and color channel information \cite{guo_depthwise_2019}, but when we added temporal information into the color channels, it could not effectively learn correlations between channels and spatial information. Despite this result, 3D CNNs which use a modified form of depthwise separable convolution have proven to be useful in efficiently capturing temporal information in video data \cite{Xie_ECCV_2018}. Lastly, it is worth noting that for the models where lag images did improve performance, the initialized model weights pre-trained on ImageNet were sufficient for doing so. 

We selected the CNN model based on its capability to generalize to visually distinct test data and prioritized temporal consistency by evaluating on TC-error. However, all CNN models occasionally displayed inconsistent classification for adjacent frames identified by expert curators as nearly identical, highlighting a lack of temporal consistency. To some degree this is expected because the CNN models have no or very little temporal information. On the other hand, because these frames are nearly identical to the human eye, this failure is more indicative of a lack of overall generalizability and is largely independent of time. These observations highlight the utility of temporal consistency in building models that can generalize well to a variety of scenarios. With the high temporal resolution of our video, we can see that subtle differences between frames can have a major impact on the model's predictions. While the factors that impact model performance are complex and varied, evaluating the temporal consistency of non-temporal models may help reveal vulnerabilities to alterations in input data caused by noise or adversarial attacks. 

Using LightGBM as the second stage of WhACC grants us the flexibility to integrate new features to improve the overall performance. For example, the Janelia whisker tracker extracts many useful features like whisker follicle angle, angular velocity, approximate distance to pole and whisker curvature. These and other useful features could be directly combined with our extracted features to improve overall performance. Alternatively, we could train a separate model on these additional features and fuse those predictions with WhACC. For example, tracking data is highly reliable at defining non-touch periods when the whisker is not near the pole and WhACC excels at classifying touch and near-touch frames. 

CNNs and other deep learning models have been combined with gradient boosting machines to achieve excellent results across multiple domains. These hybrid models have been used for image classification \cite{Ismail_Gradient_2022, sugiharti_integration_2022}, as well as time series data, including stock price forecasting \cite{Liu_GradientBoost_2019}, ultra-short-term wind power prediction \cite{ju_model_2019}, acoustic scene classification \cite{fonseca_acoustic_2017}, and sleep stage classification \cite{chambon_deep_2018}. Techniques for combining these models include using a CNN as a feature extractor and then feeding these features into a gradient-boosted machine model, which is similar to the process we employed here. Another option is to use a late-fusion approach, where models are trained in parallel and predictions are combined either by averaging or by using an additional classifier to generate the final predictions. CNNs and gradient-boosted machines have even been integrated into a unified framework \cite{thongsuwan_convxgb:_2021}. 

A study related to our work \cite{bober-irizar_learning_2019} identified precise temporal events in video data and also leveraged a gradient boosted machine but otherwise approached the problem differently. They use a late fusion approach where we use model stacking. Most importantly, they integrate temporal information using an LSTM, but we use feature engineering and allow LightGBM to capture temporal relationships. To our knowledge, no other study has combined a CNN to extract spatial features from video frames and then used feature engineering combined with a gradient-boosted machine to capture temporal information, to identify precise temporal events in video. In conclusion, WhACC is an efficient and adaptable classification software that can greatly reduce human curations hours across different laboratories which use a similar experimental design. Our package, data card, model card, and full walkthrough can be found at \url{https://github.com/hireslab/whacc}. 

\section{Materials and Methods}
\subsection{Data selection and preprocessing}
The original training data consisted of grayscale MP4 video from eight mice across eight different behavioral sessions, with two sessions each from four different scientists. The videos were collected using three different experimental rigs and across two different laboratories. We created a Google Colab compatible interface to select a template image of the pole. Using this interface we selected a template image of the pole (default 61X61 pixels). Any template size can be used to match the specific object. Using this template image we match and extract each frame in an entire session using OpenCV \cite{bradski_opencv_2000}. All data are stored in H5 files using the H5py package \cite{collette_h5py_2020}.
 
For each CNN model, training and validation videos were split based on the segments created from the 80-border extraction method in Figure 3A and described above. Test data consisted of a set from another laboratory. We reasoned that if the CNN model could perform well on these visually different images of the same whisker task, then the extracted features would be general features of the task and not overfit to the specific video conditions.
 
For each of these datasets we selected a subset using 3-border extraction, made ten copies of these data, and then augmented each frame independently prior to training. Examples of fully augmented images are displayed in Figure 1B. Using keras from tensorflow, images were randomly augmented using all of the following augmentations; full rotation, shifting in any direction up to 10\%, symmetrically zooming in or out up to 25\%, changes in brightness ranging from 20\% to 120\% of the initial value. Additionally, we used the imgaug \cite{jung_imgaug_2017} python package to modify images with additive gaussian noise, where scale was set to three. These parameters were selected based on trial and error to ensure frames were still interpretable to a human curator. For lag images we simply stacked images from the previous two time points into the different color channels.
 
\subsection{Training CNNs}
Each CNN model was initialized using the weights pre-trained on ImageNet using Tensorflow \cite{abadi_tensorflow:_2016}. Input images are fed into the network using a batch generator and automatically normalized from -1 to 1 and resized to 96X96 pixels to match pre-trained CNN model formats. Multiple levels of unfreezing were tested but only fully unfrozen models are presented here, as they performed best. We used RMSprop as the optimization algorithm and binary cross-entropy as the loss function. Various batch sizes and learning rates were also tested but final models were trained using a batch size of 200 with a learning rate of 10-6. We trained with a dropout rate of 50\% to improve generalizability of the model. We performed early stopping based on the validation loss with a patience of 15 epochs. For all models, training was stopped by early stopping. All CNN models have a final dense layer with a sigmoid activation function for binary classification. The model epoch with the best validation loss was saved and later compared against other CNN models evaluated on test data. ResNet50V2 with lag images training on augmented data was selected as the feature extractor model based on the test set having the lowest TC-error. All models which utilized lag images predicted the leading frame such that input images were at timesteps -2, -1 and 0, and the target value was at 0. 

\subsection{LSTM models}
All LSTM models were constructed with a batch size of 210 and the 2,048 input features from Resnet50V2 were Z-scored and clipped to a range of  -1 to 1, this method performed better than min-max normalization (not shown). The Resnet50V2 base model was trained on augmented images with lag. However, for training the LSTM models, only the original lag images were used to extract features through Resnet50V2. This approach was chosen to enable a direct comparison with the LightGBM model, as they were both trained under similar conditions due to memory limitations. The features were inputted into each model using timesteps at times  7 to 7 to predict the target value at time zero. Optimization was carried out using the Adam optimizer with a learning rate of 10-5. Training was regulated with early stopping based on validation loss (binary cross-entropy) with a patience of 12 epochs; the model with the best validation loss was selected and then TC-error was evaluated. The single-layer LSTM model comprises one LSTM layer with 128 units, followed by a dense layer with a sigmoid activation function for binary classification (1,114,753 total parameters). Conversely, the multi-layer LSTM model has three LSTM layers with 128, 64, and 32 units respectively, interspersed with dropout layers at a rate of 20\%, and culminates in a dense layer with a sigmoid activation function for binary classification (1,176,481 total parameters). All intermediate LSTM layers used a tanh activation function.

\subsection{Feature engineering selection}
As depicted in Figure 4, we applied various transformations to the 2,048 features extracted from the penultimate layer of ResNet50V2. These included the following. Shifting by -5, -4, -3, -2, -1, 1, 2, 3, 4, 5. Rolling mean and rolling standard deviations centered at time of time = 0 with window sizes of 3, 7, 11, 15, 21, 41, 61. Discrete difference between time = 0 and relative step sizes -50, -20, -10, -5, -4, -3, -2, -1, 1, 2, 3, 4, 5, 10, 20, 50. With the original features this was a total of 41 sets of 2,048 features. Lastly, we took each one of those sets and calculated the standard deviation across feature space to generate 41 additional features for a total of 41*2048+41 or 84,009 features.
 
For feature selection we split all available data into training (70\%) and validation (30\%) sets and then split both of those into 10 equal sets. This was necessary because our full dataset with 84,009 features could not fit into memory on our local machine with 128GB of RAM. Using an ensemble of 10 models also helped reduce the risk of selecting poor predictors due to chance. We split data based on frame index (as opposed to segment or video), to ensure variance was equally distributed across the different splits. Equal variance among these splits meant that important and reliable features should have approximately the same importance across models. This helped us to confidently eliminate features with moderate importance in a few models with little to no importance in the other models.
 
Data consisted of features generated from lag images without any augmentation from the original dataset (80-border extraction). Augmented images were not included in any LightGBM models because preliminary tests showed decreased performance when trained on augmented images. Due to the nature of our feature engineering operations, some data were undefined at the edges (e.g., smoothing with a window of 61, the last 30 frames are undefined). Because of this, we dropped any time points that contained undefined values prior to training, so that features with undefined regions were not eliminated due to underrepresentation. As mentioned in the main text, models were trained with the original 2,048 features then with the full 84,009 features for comparison. We evaluated model performance using mean AUC of validation sets. Across the ten models trained on the full feature set, only 28,913 features were used at all, so we eliminated the others. We continued to eliminate features based on threshold of gain and split importance. This process was done through trial and error and was carried out by exploring distributions of feature importance and defining thresholds to eliminate weak predictors. Naturally, some feature elimination steps decreased performance and so the actual selection process resulted in many offshoots not shown in Figure 4G. If eliminating features reduced performance, we simply came up with a more conservative elimination step and tried again. We continued this process until we could no longer maintain the same level of performance when reducing features. Once finished, we selected 2,105 features which we proceeded to use to train the LightGBm classifier. 

The LightGBM models used for feature selection were trained with the following hyperparameters. The number of leaves in the decision tree was set to 31, and the maximum number of iterations was set to 5000 but all models trained were halted before reaching this limit. The AUC was used as the evaluation metric, and early stopping was applied after 40 rounds. The histogram pool size was set to 16,384, and the maximum bin value was set to 255. The learning rate was set to 0.1, and the maximum depth of the tree was unlimited. The minimum number of data points allowed in a leaf was set to 20, and both the bagging fraction and feature fraction were set to 1. The minimum data points allowed in a bin was set to 3.

\subsection{Preprocessing and training}
ResNet50V2 was selected as our feature extractor model, bases on the performance on the original test set described in the data selection and preprocessing section summarized in Figure 3B and Table 1. For training the final WhACC model, we created different training (263,069 frames), validation (73,909 frames), and test (114,051 frames) sets by randomly assigned data from single videos. Training and validation sets were trimmed using 80-border extraction but the test set included all frames to mimic the distribution of actual data. These data were used to train the final WhACC model and for the later retraining process. Just like the feature selection step, features were generated from lag images without augmentation. Also note, the feature extractor CNN was selected based on the original test set described in the data selection and preprocessing section summarized in Figure 3B and Table 1.
 
Initial attempts to train our model showed great performance but we observed edge effects for input data within a few frames of the last frame in a video. We discovered this was due to the model relying on features which were unavailable for those timepoints due to how they were engineered. To remedy this, we created an index matrix (100 by 2,105) of all the possible data points which could contain undefined values from edge effects, for each feature at each timepoint. Next, we created a duplicate of each dataset. For each timepoint in these duplicates, we randomly selected from the 100 possible timepoints which contained undefined values and added undefined values to these relevant features to act as a mask. These altered datasets simulate what features look like for the first and last 50 frames of a video where undefined values are possible. Final datasets for the LightGBM portion of WhACC were composed of these altered datasets combined with their unaltered counterparts. This process is carried out by default when retraining WhACC on sample data as well. 

To optimize the performance of the LightGBM classifier for WhACC, we used early stopping with a custom callback that evaluated TC-errors and AUC. If the validation data failed to improve for either of these metrics for 500 rounds, training was halted, and the model with the best TC-error was selected. The number of iterations was set to 10,000 but all models were halted before reaching this limit. To determine the optimal hyperparameters, we used Optuna, a hyperparameter optimization framework. Optuna settings included L1 and L2 regularization values between $1*10^{-5}$ and 10. The number of leaves in the decision tree ranged from 2 to 256, and the suggested fraction of features to use per tree ranged from 0.4 to 1.0. The suggested fraction of data points to use for each bagging sample ranged from 0.4 to 1.0, and the suggested frequency of bagging ranged from 1 to 7. Finally, the suggested minimum number of data points allowed in a leaf was set to range from 5 to 100. 

\subsection{Retraining procedure}
To make WhACC more reliable and generalize across variability of new datasets, we designed and tested a retraining process using a small subset of data. We tested different methods for sampling data, including using a few full length videos and a few frames across many different videos. The latter was more effective. For the retraining procedure we used the object tracking location to select trials equally spaced along the horizontal coordinates of the video. Then we selected a starting frame where the pole was available in all videos. We sampled 10 frames from either 10 or 100 videos for a total of either 100 or 1,000 frames from each session respectively. Once data was selected, we used a GUI to curate the subset and saved our sample data. We then load the original training and validation data and split the new sample data derived from each session into each. We split sample data into 70/30 training and validation sets, and apply a weight of two to new data to bias the model slightly. Next we duplicate each dataset and add undefined values as described in the preprocessing and training section. Then we retrain just the LightGBM stage of WhACC using the preselected hyperparameters or by using Optuna to select them again. Finally, we evaluate each model's validation TC-error and AUC to help select a final model. 

We evaluated WhACC before and after retraining using a holdout dataset of 16 sessions.
Each session was composed of 100 videos and each video had 3,000 frames. Accounting for pole available times, we tested a total of $\sim$ 4 million frames which were not used in any stage of developing WhACC. This dataset was curated by only one expert curator, so the human error rate is not known. We compared WhACC before retraining, WhACC after retraining on an additional 100 frames from each session (1,600 total) and WhACC after training on an additional 1,000 from each session (16,000 total). We constructed PSTHs by taking the mean spikes response across all touch onsets times for either each model and the expert curator. For the expert curator 95\% confidence intervals are included (1.96* standard error). 

Touch neurons were identified using PSTH traces smoothed with a window of 5ms for visibility. Using a baseline period from -100 to -20 ms from touch onset, we subtracted the signal from the baseline and divided it by the SD of the baseline to get a Z-scored trace. Any neuron with points outside of 2 SD from the baseline for four or more continuous time points from touch onset to 100 ms after was considered a touch neuron. If a touch signal was bipolar (i.e. positively and negatively modulated at different times following touch), only the significant times directly following touch onset were considered the signal window. If two significant regions were separated by some points below two standard deviations, then the signal window was considered to be all the time points from the beginning of the first significant region to the end of the last significant region. Spikes per touch were calculated by taking the integral of the significant region of the unsmoothed baseline subtracted PSTH. We then took the absolute value of this number. 

\subsection{Evaluating performance}
We created a class to determine touch count and edge errors and determine their lengths. As explained in the main text, we utilized TC-errors to help us select a model that performed more similarly to human curators, specifically in terms of the types of errors that were deemed acceptable. We determined human error rate by comparing consensus frames between two curators against the predictions of the remaining curator, for each combination of the three curators. This method allowed for a fair comparison between WhACC and each curator. For all models we median smoothed predictions with a five ms window because we found this uniformly increased performance across models. Median smoothing primarily helped to eliminate noisy predictions found near touch onset and offset events.

%




\bibliography{WhACC_BIBLIOGRAPHY_V3}

\end{document}